\documentclass{article}

\usepackage{times}
\usepackage{graphicx} 
\usepackage{subfigure} 

\usepackage{natbib}

\usepackage{algorithm}
\usepackage{algorithmic}

\usepackage{hyperref}

\usepackage[accepted]{icml2017}

\usepackage{amsmath}
\usepackage{amssymb}
\usepackage{bm}

\DeclareMathOperator*{\argmax}{argmax}

\icmltitlerunning{Interactive Learning from Policy-Dependent Human Feedback}

\begin{document} 

\twocolumn[
\icmltitle{Interactive Learning from Policy-Dependent Human Feedback}



\icmlsetsymbol{equal}{*}

\begin{icmlauthorlist}
\icmlauthor{James MacGlashan}{alpha}
\icmlauthor{Mark K Ho}{beta}
\icmlauthor{Robert Loftin}{delta}
\icmlauthor{Bei Peng}{gamma}
\icmlauthor{Guan Wang}{beta}
\icmlauthor{David L. Roberts}{delta} 
\icmlauthor{Matthew E. Taylor}{gamma}
\icmlauthor{Michael L. Littman}{beta}
\end{icmlauthorlist}

\icmlaffiliation{alpha}{Cogitai}
\icmlaffiliation{beta}{Brown University}
\icmlaffiliation{delta}{North Carolina State University}
\icmlaffiliation{gamma}{Washington State University}

\icmlcorrespondingauthor{James MacGlashan}{james@cogitai.com}

\icmlkeywords{reinforcement learning, human-robot interaction}

\vskip 0.3in
]



\printAffiliationsAndNotice{\icmlEqualContribution} 

\begin{abstract} 
This paper investigates the problem of interactively learning behaviors communicated by a human teacher using positive and negative feedback. Much previous work on this problem has made the assumption that people provide feedback for decisions that is dependent on the behavior they are teaching and is independent from the learner's current policy. We present empirical results that show this assumption to be false---whether human trainers give a positive or negative feedback for a decision is influenced by the learner's current policy. 
Based on this insight, we introduce {\em Convergent Actor-Critic by Humans} (COACH), an algorithm for learning from policy-dependent feedback that converges to a local optimum. Finally, we demonstrate that COACH can successfully learn multiple behaviors on a physical robot.
\end{abstract}

\section{Introduction}

Programming robots is very difficult, in part because the real world is inherently rich and---to some degree---unpredictable. In addition, our expectations for physical agents are quite high and often difficult to articulate. Nevertheless, for robots to have a significant impact on the lives of individuals, even non-programmers need to be able to specify and customize behavior.
Because of these complexities, relying on end-users to provide instructions to robots programmatically seems destined to fail. 

Reinforcement learning (RL) from human trainer feedback provides a compelling alternative to programming because agents can learn complex behavior from very simple positive and negative signals. Furthermore, real-world animal training is an existence proof that people can train complex behavior using these simple signals. Indeed, animals have been successfully trained to guide the blind, locate mines in the ocean, detect cancer or explosives, and even solve complex, multi-stage puzzles. 

Despite success when learning from environmental reward, traditional reinforcement-learning algorithms have yielded limited success when the reward signal is provided by humans.
This failure underscores the importance that algorithms for learning from humans are based on appropriate models of human-feedback. Indeed, much human-centered RL work has investigated and employed different models of human-feedback~\cite{knox2009interactively,thomaz2006reinforcement,thomaz2007robot,thomaz08,griffith2013policy,loftin15}.
Many of these 
algorithms leverage the observation that people tend to give feedback that is best interpreted as guidance on the policy the agent should be following, rather than as a numeric value to be maximized by the agent. However, these approaches assume models of feedback that are independent of the policy the agent is currently following. We present empirical results that demonstrate that this assumption is incorrect and further demonstrate cases in which policy-independent learning algorithms suffer from this assumption. Following this result, we present {\em Convergent Actor-Critic by Humans} (COACH), an algorithm for learning from policy-dependent human feedback. COACH is based on the insight that the {\em advantage function} (a value roughly corresponding to how much better or worse an action is compared to the current policy) provides a better model of human feedback, capturing human-feedback properties like diminishing returns, rewarding improvement, and giving 0-valued feedback a semantic meaning that combats forgetting. We compare COACH to other approaches in a simple domain with simulated feedback. Then, to validate that COACH scales to complex problems, we train five different behaviors on a TurtleBot robot. 


\section{Background}

For modeling the underlying decision-making problem of an agent being taught by a human, we adopt the Markov Decision Process (MDP) formalism. An MDP is a 5-tuple: $\langle S, A ,T, R, \gamma \rangle$, where $S$ is the set of possible states of the environment; $A$ is the set of actions available to the agent; $T(s' | s, a)$ is the transition function, which defines the probability of the environment transitioning to state $s'$ when the agent takes action $a$ in environment state $s$; $R(s, a, s')$ is the reward function specifying the numeric reward the agent receives for taking action $a$ in state $s$ and transitioning to state $s'$; and $\gamma \in [0, 1]$ is a discount factor specifying how much immediate rewards are preferred to more distant rewards.

A {\em stochastic policy} $\pi$ for an MDP is a per-state action probability distribution that defines an agent's behavior; $\pi : S \times A \rightarrow [0, 1]$, where $\sum_{a \in A} \pi(s,a) = 1, \forall s \in S$. In the MDP setting, the goal is to find the optimal policy $\pi^*$, which maximizes the expected future discounted reward when the agent selects actions in each state according to $\pi^*$; $\pi^* = \argmax_\pi E[\sum_{t=0}^\infty \gamma^t r_t | \pi]$, where $r_t$ is the reward received at time $t$. Two important concepts in MDPs are the value function ($V^\pi$) and action--value function ($Q^\pi$). The value function defines the expected future discounted reward from each state when following some policy 
and the action--value function defines the expected future discounted reward when an agent takes some action in some state and then follows some policy $\pi$ thereafter.
These equations can be recursively defined via the Bellman equation: $V^\pi(s) = \sum_a \pi(s,a) Q^\pi(s, a)$ and $Q^\pi(s, a) = \sum_{s'} T(s' | s, a) \left[ R(s, a, s') + \gamma V^\pi(s')\right]$. For shorthand, the value functions for the optimal policies are usually denoted $V^*$ and $Q^*$.

In reinforcement learning (RL), an agent interacts with an environment modeled as an MDP, but does not have direct access to the transition function or reward function and instead must learn a policy from environment observations. A common class of RL algorithms are {\em actor-critic} algorithms. \citet{bhatnagar2009natural} provide a general template for these algorithms. Actor-critic algorithms are named for the two main components of the algorithms: The actor is a parameterized policy that dictates how the agent selects actions; the critic estimates the value function for the actor and provides critiques at each time step that are used to update the policy parameters. Typically, the critique is the temporal difference (TD) error: $\delta_t = r_t + \gamma V(s_t) - V(s_{t-1})$, which describes how much better or worse a transition went than expected. 

\section{Human-centered Reinforcement Learning}
In this work, a {\em human-centered reinforcement-learning} (HCRL) problem is a learning problem in which an agent is situated in an environment described by an MDP but in which rewards are generated by a human trainer instead of from a stationary MDP reward function that the agent is meant to maximize.  The trainer has a target policy $\pi^*$ they are trying to teach the agent. The trainer communicates this policy by giving numeric feedback as the agent acts in the environment.
The goal of the agent is to learn the target policy $\pi^*$ from the feedback.

To define a learning algorithm for this problem, we first characterize how human trainers typically use numeric feedback to teach target policies. If feedback is stationary and intended to be maximized, it can be treated as a reward function and standard RL algorithms used.
Although this approach has had some success~\cite{pilarski2011online,isbell2001social}, there are complications that limit its applicability. In particular, a trainer must take care that the feedback they give contains no unanticipated exploits, constraining the feedback strategies they can use. Indeed, prior research has shown that interpreting human feedback like a reward function often induces {\em positive reward cycles} that lead to unintended behaviors~\cite{knoxthesis,ho15}. 

The issues with interpreting feedback as reward have led to the insight that human feedback is better interpreted as commentary on the agent's behavior; for example, positive feedback roughly corresponds to ``that was good'' and negative feedback roughly corresponds to ``that was bad.'' In the next section, we review existing HCRL approaches that build on this insight. 

\section{Related Work}
A number of existing approaches to HCRL and RL that includes human feedback has been explored in the past. The most similar to ours, and a primary inspiration for this work, is the TAMER framework~\cite{knoxthesis}.
In TAMER, trainers provide interactive numeric feedback as the learner takes actions. The learner attempts to estimate a target reward function by interpreting trainer feedback as exemplars of this function. When the agent makes rapid decisions, TAMER divides the feedback among the recent state--action pairs according to a probability distribution. TAMER makes decisions by myopically choosing the action with the highest reward estimate. Because the agent myopically maximizes reward, the feedback can also be thought of as exemplars of $Q^*$. Later work also investigated non-myopically maximizing the learned reward function with a planning algorithm~\cite{knox2013learning}, but this approach requires a model of the environment
and special treatment of termination conditions.

Two other closely related approaches are SABL~\cite{loftin15} and Policy Shaping~\cite{griffith2013policy}. Both of these approaches treat feedback as discrete probabilistic evidence of the trainer's target parameterized policy. SABL's probabilistic model additionally includes (learnable) parameters for describing how often a trainer is expected to give explicit positive or negative feedback.

There have also been some domains in which treating human feedback as reward signals to maximize has had some success, such as in shaping the control for a prosthetic arm~\cite{pilarski2011online} and learning how to interact in an online chat room from multiple users' feedback~\cite{isbell2001social}. Some complications with how people give feedback have been reported, however.

Some research has also examined combining human feedback with more traditional environmental rewards~\cite{AAMAS10-knox,tenorio2010dynamic,clouse1992teaching,maclin2005giving}. A challenge in this context in practice is that rewards do not naturally come from the environment and must be programmatically defined. However, it is appealing because the agent can learn in the absence of an active trainer. We believe our approach to HCRL could also straightforwardly incorporate learning from environmental reward as well, but we leave this investigation for future work.

Finally, a related research area is {\em learning from demonstration} (LfD), in which a human provides examples of the desired behavior. There are a number of different approaches to solving this problem surveyed by \citet{argall2009survey}. We see these approaches as complementary to HCRL because it is not always possible, or convenient, to provide demonstrations. LfD approaches that learn a parameterized policy could also operate with COACH, allowing the agent to have their policy seeded by demonstrations, and then fine tuned with interactive feedback.

Note that the policy-dependent feedback we study here is viewed as essential in behavior analysis reinforcement schedules~\cite{miltenberger2011behavior}. Trainers are taught to provide diminishing returns (gradual decreases in positive feedback for good actions as the agent adopts those actions), differential feedback (varied magnitude of feedbacks depending on the degree of improvement or deterioration in behavior), and policy shaping (positive feedback for suboptimal actions that improve behavior and then negative feedback after the improvement has been made), all of which are policy dependent.

\section{Policy-dependent Feedback}

A common assumption of existing HCRL algorithms is that feedback depends only on the quality of an agent's action selection. An alternative hypothesis is that feedback also depends on the agent's current policy. That is, an action selection may be more greatly rewarded or punished depending on how often the agent would typically be inclined to select it. For example, more greatly rewarding the agent for improving its performance than maintaining the status quo. We call the former model of feedback {\em policy-independent} and the latter {\em policy-dependent}. If people are more naturally inclined toward one model of feedback, algorithms based on the wrong assumption may result in unexpected responses to feedback. Consequently, we were interested in investigating which model better fits human feedback.

Despite existing HCRL algorithms assuming policy-independent feedback, evidence of policy-dependent feedback can be found in prior works with these algorithms. For example, it was often observed that trainers taper their feedback over the course of learning~\cite{ho15,knox2012humans,isbell2001social}. 
Although diminishing feedback is a property that is explained by people's feedback being policy-dependent---as the learner's performance improves, trainer feedback is decreased---an alternative explanation is simply trainer fatigue. To further make the case for human feedback being policy dependent, we provide a stronger result showing that 
trainers---for the same state--action pair---choose positive or negative feedback depending on their perception of the learner's behavior. 

\subsection{Empirical Results}
\begin{figure}
\centering
\includegraphics[width=0.8\columnwidth]{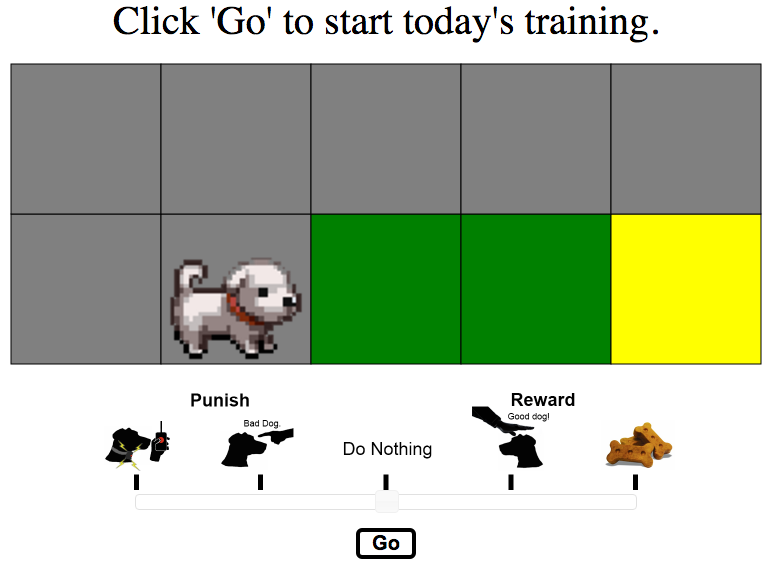}
\caption{The training interface shown to AMT users.
\vspace{-10pt}
}
\label{fig:interface}
\end{figure}

We had Amazon Mechanical Turk (AMT) participants teach an agent in a simple sequential task, illustrated
in Figure~\ref{fig:interface}. Participants were instructed to train a virtual dog to walk to the yellow goal location in a grid world as fast as possible but without going through the green cells. They were additionally told that, as a result of prior training, their dog was already either ``bad,'' ``alright,'' or ``good'' at the task and were shown examples of each behavior before training. In all cases, the dog would start in the location shown in Figure~\ref{fig:interface}. ``Bad'' dogs walked straight through the green cells to the yellow cell. ``Alright'' dogs first moved left, then up, and then to the goal, avoiding green but not taking the shortest route. ``Good'' dogs took the shortest path to yellow without going through green. 

During training, participants saw the dog take an action from one tile to another and then gave feedback after every action using a continuous labeled slider as shown. The slider always started in the middle of the scale on each trial, and several points were labeled with different levels of reward (praise and treats) and punishment (scolding and a mild electric shock). Participants went through a brief tutorial using this interface. Responses were coded as a numeric value from $-50$ to $50$, with ``Do Nothing'' as the zero-point.

During the training phase, participants trained a dog for three {\em episodes} that all started in the same position and ended at the goal. The dog's behavior was pre-programmed in such a way that the first step of the final episode would reveal if feedback was policy dependent. Each user was placed into one of three different conditions: improving, steady, or degrading. For all three conditions, the dog's behavior in the final episode was ``alright,'' regardless of any prior feedback. The conditions differed in terms of the behavior users observed in the first two episodes. In the first two episodes, users observed bad behavior in the improving condition (improving to alright); alright behavior in the steady condition; and good behavior in the degrading condition.
If feedback is policy-dependent, we would expect more positive feedback in the final episode for the improving condition, but not for policy-independent feedback since it was the same final behavior for all conditions.

\begin{figure}
\centering
\includegraphics[width=0.8\columnwidth]{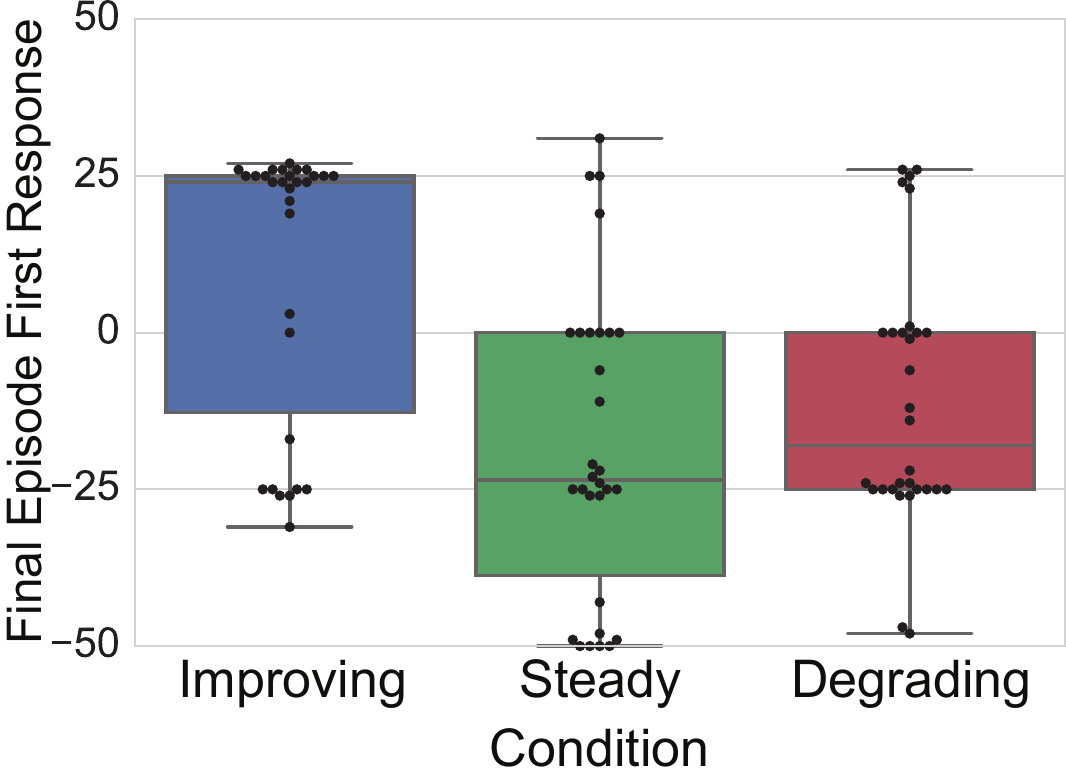}
\caption{The feedback distribution for first step of the final episode for each condition. Feedback tended to be positive for improving behavior, but negative otherwise.}
\label{fig:feedDist}
\end{figure}

Figure~\ref{fig:feedDist} shows boxplots and individual responses for the first step of the final episode under each of the three conditions. These results indicate that the sign of feedback is sensitive to the learner's policy, as predicted. The mean and median feedback under the improving condition is slightly positive (Mean = $9.8$, Median = $24$, S.D. = $22.2$; planned Wilcoxon one-sided signed-rank test: $Z = 1.71, p < 0.05$), whereas it is negative for the steady condition (Mean = $-18.3$, Median = $-23.5$, S.D. = $24.6$; planned Wilcoxon two-sided signed-rank test: $Z = -3.15, p < 0.01$) and degrading condition (Mean = $-10.8$, Median = $-18.0$, S.D. = $20.7$; planned Wilcoxon one-sided signed-rank test: $Z = -2.33, p < 0.05$). There was a main effect across the three conditions ($p < 0.01$, Kruskal-Wallace Test), and pairwise comparisons indicated that only the improving condition differed from steady and degrading conditions ($p < 0.01$ for both, Bonferroni-corrected, Mann-Whitney Pairwise test). 

\section{Convergent Actor-Critic by Humans}

In this section, we introduce {\em Convergent Actor-Critic by Humans} (COACH), an actor-critic-based algorithm capable of learning from policy-dependent feedback. COACH is based on the insight that the advantage function is a good model of human feedback and that actor--critic algorithms update a policy using the critic's TD error, which is an unbiased estimate of the advantage function. Consequently, an agent's policy can be directly modified by human feedback without a critic component. We first define the advantage function and its interpretation as trainer feedback. Then, we present the general update rule for COACH and its convergence. Finally, we present {\em Real-time COACH}, which includes mechanisms for providing variable magnitude feedback and learning in problems with a high-frequency decision cycle.

\subsection{The Advantage Function and Feedback}

The advantage function~\cite{baird1995residual} $A^\pi$ is defined as
\begin{equation}
A^\pi(s, a) = Q^\pi(s, a) - V^\pi(s).
\end{equation} 
Roughly speaking, the advantage function describes how much better or worse an action selection is compared to the agent's performance under policy $\pi$. The function is closely related to the update used in policy iteration~\cite{Puterman94}: defining $\pi'(s) = \argmax_a A^\pi(s,a)$ is guaranteed to produce an improvement over $\pi$ whenever $\pi$ is suboptimal. It can also be used in policy gradient methods to gradually improve the performance of a policy, as described later.

It is worth nothing that feedback produced by the advantage function is consistent with that recommended in behavior analysis. It trivially results in differential feedback since it is defined as the magnitude of improvement of an action over its current policy. It induces diminishing returns because, as $\pi$ improves opportunities to improve on it decrease. Indeed, once $\pi$ is optimal, all advantage-function-based feedback is zero or negative. Finally, advantage function feedback induces policy shaping in that whether feedback is positive or negative for an action depends on whether it is a net improvement over the current behavior.

\subsection{Convergence and Update Rule}

Given a performance metric $\rho$, \citet{sutton1999policy} derive a policy gradient algorithm of the form:
$
\Delta \bm{\theta} = \alpha \nabla_{\bm \theta} \rho
$.
Here, ${\bm \theta}$ represents the parameters that control the agent's behavior and $\alpha$ is a learning rate. 
Under the assumption that $\rho$ is the discounted expected reward from a fixed start state distribution, they show that
$$
\nabla_{\bm \theta} \rho = \sum_s d^\pi(s) \sum_a \nabla_{\bm \theta} \pi(s,a) Q^\pi(s,a),
$$
where $d^\pi(s)$ is the component of the (discounted) stationary distribution at $s$.
A benefit of this form of the gradient is that, given that states are visited according to $d^\pi(s)$ and actions are taken according to $\pi(s,a)$, the update at time $t$ can be made as:
\begin{equation}
\Delta {\bm \theta}_t = \alpha_t \nabla_{\bm \theta} \pi(s_t,a_t) \frac{f_{t+1}}{\pi(s_t,a_t)},
\label{e:update}
\end{equation}
where $E[f_{t+1}] = Q^\pi(s_t,a_t) - v(s)$ for any action-independent function $v(s)$.



In the context of the present paper, $f_{t+1}$ represents the feedback provided by the trainer. It follows trivially that if the trainer chooses the policy-dependent feedback $f_t = Q^\pi(s_t,a_t)$, we obtain a convergent learning algorithm that (locally) maximizes discounted expected reward. In addition, feedback of the form $f_t = Q^\pi(s_t,a_t)-V^\pi(s_t) = A^\pi(s_t, a_t)$ also results in convergence.
Note that for the trainer to provide feedback in the form of $Q^\pi$ or $A^\pi$, they would need to ``peer inside'' the learner and observe its policy. In practice, the trainer estimates $\pi$ by observing the agent's actions.



\subsection{Real-time COACH}

There are challenges in implementing Equation~\ref{e:update} for real-time use in practice. Specifically, the interface for providing variable magnitude feedback needs to be addressed, and the question of how to handle sparseness and the timing of feedback needs to be answered. Here, we introduce {\em Real-time COACH}, shown in Algorithm~\ref{alg:coach}, to address these issues.

For providing variable magnitude reward, we use reward aggregation~\cite{knox2009interactively}. In reward aggregation, a trainer selects from a discrete set of feedback values and further raises or lowers the numeric value by giving multiple feedbacks in succession that are summed together.

While sparse feedback is not especially problematic (because no feedback results in no change in policy), it may slow down learning unless the trainer is provided with a mechanism to allow feedback to affect a history of actions. We use {\em eligibility traces}~\cite{6313077} to help apply feedback to the relevant transitions. An eligibility trace is a vector that keeps track of the policy gradient and decays exponentially with a parameter $\lambda$. Policy parameters are then updated in the direction of the trace, allowing feedback to affect earlier decisions.
However, a trainer may not always want to influence a long history of actions. Consequently, Real-time COACH maintains multiple eligibility traces with different temporal decay rates and the trainer chooses which eligibility trace to use for each update. This trace choice may be handled implicitly with the feedback value selection or explicitly.


Due to reaction time, human feedback is typically delayed by about $0.2$ to $0.8$ seconds from the event to which they meant to give feedback~\cite{knoxthesis}. To handle this delay, feedback in Real-time COACH is associated with events from $d$ steps ago to cover the gap. Eligibility traces further smooth the feedback to older events.

Finally, we note that just as there are numerous variants of actor-critic update rules, similar variations can be used in the context of COACH.

\begin{algorithm}[t]
\begin{algorithmic}
\REQUIRE{policy $\pi_{\bm{\theta}_0}$, trace set $\bm{\lambda}$, delay $d$, learning rate $\alpha$}
\STATE Initialize traces $\bm{e}_\lambda \gets \bm{0} \mbox{ } \forall \lambda \in \bm{\lambda}$ 
\STATE observe initial state $s_0$
\FOR{$t=0 \text{ to } \infty$}
\STATE select and execute action $a_t \sim \pi_{\bm{\theta}_t}(s_t, \cdot)$
\STATE observe next state $s_{t+1}$, sum feedback $f_{t+1}$, and $\lambda$
\FOR{$\lambda' \in \bm{\lambda}$}
\STATE $\bm{e}_{\lambda'} \gets \lambda' \bm{e}_{\lambda'} + \frac{1}{\pi_{\bm{\theta}_t}(s_{t-d}, a_{t-d})}\nabla_{\bm{\theta}_t}\pi_{\bm{\theta}_t}(s_{t-d}, a_{t-d})$
\ENDFOR
\STATE $\bm{\theta}_{t+1} \gets \bm{\theta}_t + \alpha f_{t+1} \bm{e}_\lambda$
\ENDFOR
\end{algorithmic}
\caption{Real-time COACH}
\label{alg:coach}
\end{algorithm}

\section{Comparison of Update Rules}

To understand the behavior of COACH under different types of trainer feedback strategies, we carried out a controlled comparison in a simple grid world.
The domain is essentially an expanded version of the dog domain used in our human-subject experiment. It is a $8\times 5$ grid in which the agent starts in $0,0$ and must get to $7,0$, which yields $+5$ reward. However, from $1,0$ to $6,0$ are cells the agent needs to avoid, which yield $-1$ reward.

\subsection{Learning Algorithms and Feedback Strategies}

Three types of learning algorithms were tested. Each maintains an internal data structure, which it updates with feedback of the form $\langle s, a, f, s'\rangle$, where $s$ is a state, $a$ is an action taken in that state, $f$ is the feedback received from the trainer, and $s'$ is the resulting next state. The algorithm also must produce an action for each state encountered.

The first algorithm, Q learning~\cite{watkins92}, represents a standard value-function-based RL algorithm designed for reward maximization under delayed feedback. It maintains a data structure $Q(s,a)$, initially $0$.
Its update rule has the form:
\begin{equation}
\Delta Q(s, a) = \alpha [f + \gamma \max_{a'} Q(s',a') - Q(s, a)].
\end{equation}
Actions are chosen using the rule: $\argmax_a Q(s,a)$, where ties are broken randomly. We tested a handful of parameters and used the best values: discount factor $\gamma = 0.99$ and learning rate $\alpha = 0.2$.

In TAMER~\cite{knox09}, a trainer provides interactive numeric feedback that is interpreted as an exemplar of the reward function for the demonstrated state--action pair as the learner takes actions. We assumed that each feedback applies to the last action, and thus used a simplified version of the algorithm that did not attempt to spread updates over multiple transitions. TAMER maintains a data structure $R_H(s,a)$ for the predicted reward in each state, initially $0$. It is updated by: $\Delta R_H(s,a) = \alpha f$. We used $\alpha=0.2$. Actions are chosen via an $\epsilon$-greedy rule on $R_H(s,a)$ with $\epsilon=0.2$.

Lastly, we examined COACH, which is also designed to work well with human-generated feedback. We used a softmax policy with a single $\lambda  = 0$ trace. The parameters were a matrix of values $\theta(s,a)$, initially zero. The stochastic policy defined by these parameters was
\begin{equation*}
\pi(s,a) = e^{\beta \theta(s,a)} / \sum_a e^{\beta \theta(s,a)},
\end{equation*}
with $\beta=1$. Parameters were updated via
\begin{equation}
\Delta {\bm \theta} = \alpha \nabla_{\bm \theta} \pi(s,a) \frac{f}{\pi(s,a)},
\end{equation}
where $\alpha$ is a learning rate. We used $\alpha=0.05$.
%



In effect, each of these learning rules makes an assumption about the kind of feedback it expects trainers to use. We wanted to see how they would behave with feedback strategies that matched these assumptions and those that did not. The first feedback strategy we studied is the classical task-based reward function (``task'') where the feedback is sparse: $+5$ reward when the agent reaches the goal state, $-1$ for avoidance cells, and $0$ for all other transitions. Q-learning is known to converge to optimal behavior with this type of feedback. The second strategy provides policy-independent feedback for each state--action pair (``action''): $+5$ when the agent reaches termination, $+1$ reward when the selected action matches an optimal policy, $-1$ for reaching an avoidance cell, and $0$ otherwise. This type of feedback serves TAMER well. The third strategy (``improvement'') used feedback defined by the advantage function of the learner's current policy $\pi$, $A^\pi(s, a) = Q^\pi(s, a) - V^\pi(s)$, where the value functions are defined based on the task rewards.
This type of feedback is very well suited to COACH.

\subsection{Results}

Each combination of algorithm and feedback strategy was run $99$ times with the median value of the number of steps needed to reach the goal reported. Episodes were ended after $1,000$ steps if the goal was not reached.



\newcommand{\gs}{0.48}
\begin{figure}[t!]
\centering
\begin{tabular}{c}
\hspace{-25pt}
\includegraphics[scale=\gs]{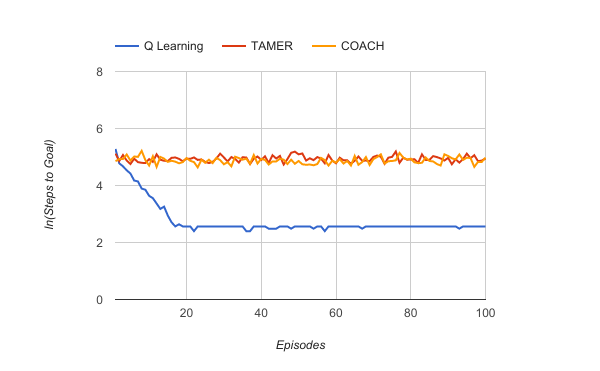} \\
(a) Task feedback \\
\hspace{-25pt}
\includegraphics[scale=\gs]{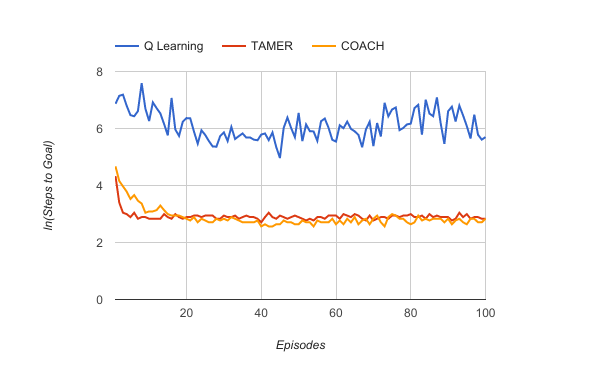}\\
(b) Action feedback \\
\hspace{-25pt}
\includegraphics[scale=\gs]{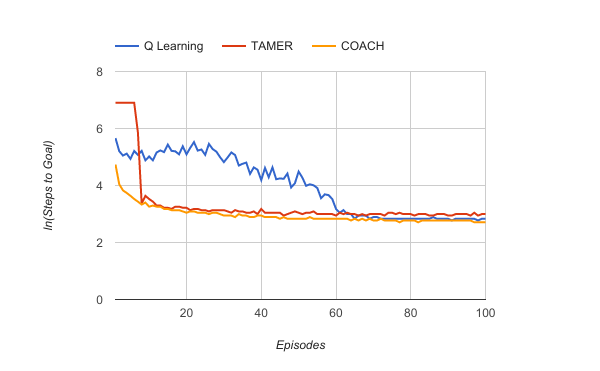}\\
(c) Improvement feedback
\end{tabular}
\caption{Steps to goal for Q learning (blue), TAMER (red), and COACH (yellow) in Cliff world under different feedback strategies. The y-axis is on a logarithmic scale.
\vspace{-10pt}
}
\label{Classical}
\end{figure}

Figure~\ref{Classical}(a) shows the steps needed to reach the goal for the three algorithms trained with task feedback. The figure shows that TAMER can fail to learn in this setting. COACH also performs poorly with $\lambda=0$, which prevents feedback from influencing earlier decisions. We did a subsequent experiment (not shown) with $\lambda=0.9$ and found that COACH converged to reasonable behavior, although not as quickly as Q learning. This result helps justify using traces to combat the challenges of delayed feedback.





Figure~\ref{Classical}(b) shows results with action feedback. This time, Q learning fails to perform well, a consequence of this feedback strategy inducing positive behavior cycles as it tries to avoid ending the trial, the same kind of problem that HCRL algorithms have been designed to avoid. Both TAMER and COACH perform well with this feedback strategy. TAMER performs slightly better than COACH, as
this is precisely the kind of feedback TAMER was designed to handle.



Figure~\ref{Classical}(c) shows the results of the three algorithms with improvement feedback, which is generated via the advantage function defined on the learner's current policy. These results tells a different story. Here, COACH performs the best. Q-learning largely flounders for most of the time, but with enough training sometimes start to converge. (Although, 14\% of the time, Q learning fails to do well even after 100 training episodes). TAMER, on the other hand, performs very badly at first. While the median score in the plot shows TAMER suddenly performing more comparably to COACH after about 10 episodes, 29\% of our training trials completely failed to improve and timed-out across all 100 episodes.

\section{Robotics Case Study}

In this section, we present qualitative results on Real-time COACH applied to a TurtleBot robot. The goal of this study was to test that COACH can scale to a complex domain involving multiple challenges, including training an agent that operates on a fast decision cycle (33ms), noisy non-Markov observations from a camera, and agent perception that is hidden from the trainer. To demonstrate the flexibility of COACH, we trained it to perform five different behaviors involving a pink ball and cylinder with an orange top using the same parameter selections. We discuss these behaviors below.  We also contrast the results to training with TAMER. We chose TAMER as a comparison because, to our knowledge, it is the only HCRL algorithm with success on a similar platform~\cite{knox2013training}.

The TurtleBot is a mobile base with two degrees of freedom that senses the world from a Kinect camera. We discretized the action space to five actions: forward, backward, rotate clockwise, rotate counterclockwise, and do nothing. The agent selects one of these actions every 33ms. To deliver feedback, we used a Nintendo Wii controller to give $+1$, $+4$, or $-1$ numeric feedback, and pause and continue training. For perception, we used only the RGB image channels from the Kinect. Because our behaviors were based around a relocatable pink ball and a fixed cylinder with an orange top, we hand constructed relevant image features to be used by the learning algorithms. These features were generated using techniques similar to those used in neural network architectures. 
The features were constructed by first transforming the image into two color channels associated with the colors of the ball and cylinder. Sum pooling to form a lower-dimensional $8\times 8$ grid was applied to each color channel. Each sum-pooling unit was then passed through three different normalized threshold units defined by $T_i(x) = \min(\frac{x}{\phi_i},1)$, where $\phi_i$ specifies the saturation point.
Using multiple saturation parameters differentiates the distance of objects, resulting in three ``depth'' scales per color channel. 
Finally, we passed these results through a $2\times 8$ max-pooling layer with stride 1. 

The five behaviors we trained were push--pull, hide, ball following, alternate, and cylinder navigation. In push--pull, the TurtleBot is trained to navigate to the ball when it is far, and back away from it when it is near. The hide behavior has the TurtleBot back away from the ball when it is near and turn away from it when it is far. In ball following, the TurtleBot is trained to navigate to the ball. In the alternate task, the TurtleBot is trained to go back and forth between the cylinder and ball. 
Finally, cylinder navigation involves the agent navigating to the cylinder. We further classify training methods for each of these behaviors as {\em flat}, involving the push--pull, hide, and ball following behaviors; and {\em compositional}, involving the alternate and cylinder navigation behaviors.

In all cases, our human trainer (one of the co-authors) used differential feedback and diminishing returns to quickly reinforce behaviors and restrict focus to the areas needing tuning. However,
in alternate and cylinder navigation, they attempted more advanced compositional training methods.
For alternate, the agent was first trained to navigate to the ball when it sees it, and then turn away when it is near. Then, the same was independently done for the cylinder. After training, introducing both objects would cause the agent to move back and forth between them. For cylinder navigation, they attempted to make use of an animal-training method called {\em lure training} in which an animal is first conditioned to follow a lure object, which is then used to guide it through more complex behaviors. In cylinder navigation, they first trained the ball to be a lure, used it to guide the TurtleBot to the cylinder, and finally gave a $+4$ reward to reinforce the behaviors it took when following the ball (turning to face the cylinder, moving toward it, and stopping upon reaching it). The agent would then navigate to the cylinder without requiring the ball to be present.


For COACH parameters, we used a softmax parameterized policy, where each action preference value was a linear function of the image features, plus $\tanh(\theta_a)$, where $\theta_a$ is a learnable parameter for action $a$, providing a preference in the absence of any stimulus. We used two eligibility traces with $\lambda=0.95$ for feedback $+1$ and $-1$, and $\lambda=0.9999$ for feedback $+4$. The feedback-action delay $d$ was set to 6, which is $0.198$ seconds. Additionally, we used an actor-critic parameter-update rule variant in which action preference values are directly modified (along its gradient), rather than by the gradient of the policy~\cite{sutton1998reinforcement}. This variant more rapidly communicates stimulus--response preferences. For TAMER, we used typical parameter values for fast decision cycle problems: delay-weighted aggregate TAMER with uniform distribution credit assignment over $0.2$ to $0.8$ seconds, $\epsilon_p=0$, and $c_{min}=1$~\cite{knoxthesis}. (See prior work for parameter meaning.) TAMER's reward-function approximation used the same representation as COACH.

\subsection{Results and Discussion}

COACH was able to successfully learn all five behaviors and a video showing its learning is available online at \url{https://www.youtube.com/watch?v=e2Ewxumy8EA}. Each of these behaviors were trained in less than two minutes, including the time spent verifying that a behavior worked.  Differential feedback and diminishing returns allowed only the behaviors in need of tuning to be quickly reinforced or extinguished without any explicit division between training and testing. Moreover, the agent successfully benefited from the compositional training methods, correctly combining subbehaviors for alternate, and quickly learning cylinder navigation with the lure. 

TAMER only successfully learned the behaviors using the flat training methodology and failed to learn the compositionally trained behaviors. In all cases, TAMER tended to forget behavior, requiring feedback for previous decisions it learned to be resupplied after it learned a new decision. For the alternate behavior, this forgetting led to failure: after training the behavior for the cylinder, the agent forgot some of the ball-related behavior and ended up drifting off course when it was time to go to the ball. TAMER also failed to learn from lure training because
TAMER does not allow reinforcing a long history of behaviors.

We believe TAMER's forgetting is  a
result of interpreting feedback as reward-function exemplars in which new feedback in similar contexts can change the target. 
To illustrate this problem, we constructed a well-defined scenario in which TAMER consistently unlearns behavior. In this scenario, the goal was for the TurtleBot to always stay whenever the ball was present, and move forward if just the cylinder was present. We first trained TAMER to stay when the ball alone was present using many rapid rewards (yielding a large aggregated signal). Next, we trained it to move forward when the cylinder alone was present. We then introduced both objects, and the TurtleBot correctly stayed. After rewarding it for \emph{staying} with a single reward (weaker than the previously-used many rapid rewards), the TurtleBot responded by moving forward---the positive feedback actually caused it to unlearn the rewarded behavior. This counter-intuitive response is a consequence of the small reward decreasing its reward-function target for the stay action to a point lower than the value for moving forward. Roughly, because TAMER does not treat zero reward as special, a positive reward can be a negative influence if it is less than expected. COACH does not exhibit this problem---any positive reward for staying will strengthen the behavior.

\section{Conclusion}
\vspace{-2pt}
In this work, we presented empirical results that show that the numeric feedback people give agents in an interactive training paradigm is influenced by the agent's current policy and argued why such policy-dependent feedback enables useful training strategies. We then introduced COACH, an algorithm that, unlike existing human-centered reinforcement-learning algorithms, converges to a local optimum when trained with policy-dependent feedback. We showed that COACH learns robustly in the face of multiple feedback strategies and finally showed that COACH can be used in the context of robotics with advanced training methods.

There are a number of exciting future directions to extend this work. In particular, because COACH is built on the actor-critic paradigm, it should be possible to combine it straightforwardly with learning from demonstration and environmental rewards, allowing an agent to be trained in a variety of ways. Second, because people give policy-dependent feedback, 
investigating how people model the current policy of the agent and how their model differs from the agent's actual policy
may produce even greater gains.

\section*{Acknowledgements}
\vspace{-2pt}
We thank the anonymous reviewers for their useful suggestions and comments. This research has taken place in part at the Intelligent Robot Learning (IRL) Lab, Washington State University. IRL's support includes NASA NNX16CD07C, NSF IIS-1149917, NSF IIS-1643614, and USDA 2014-67021-22174.

\newpage

\bibliographystyle{icml2017}
\bibliography{main}

\end{document}